\documentclass[10pt,twocolumn,letterpaper]{article}

\usepackage{cvpr}
\usepackage{times}
\usepackage{epsfig}
\usepackage{graphicx}
\usepackage{amsmath}
\usepackage{amssymb}

\usepackage[breaklinks=true,bookmarks=false]{hyperref}

\cvprfinalcopy % *** Uncomment this line for the final submission

\def\cvprPaperID{****} % *** Enter the CVPR Paper ID here

\begin{document}

%%%%%%%%% TITLE
\title{An Effective Way to Improve YouTube-8M Classification Accuracy in Google Cloud Platform
	%Ensembling Method for Youtube-8M Multi-label Classification Problem
}

\author{Zhenzhen Zhong\\
	{\tt\small zhenzhenzhongise@gmail.com}
	\and
	Shujiao Huang\\
	{\tt\small shujiao.h@gmail.com}
	\and
	Cheng Zhan\\
	{\tt\small zhancheng0511@gmail.com}
	\and
	Licheng Zhang\\
	{\tt\small xiyao.fei@gmail.com}
	\and
	Zhiwei Xiao\\
	{\tt\small zhiwei.xiao@gmail.com}
	\and
	Chang-chun Wang\\
	{\tt\small wither81@gmail.coms}
	\and
	Pei Yang\\
	{\tt\small peiyangdailylife@gmail.com}
}
\maketitle
%\thispagestyle{empty}

%%%%%%%%% ABSTRACT
\begin{abstract}
	
	Large-scale datasets have played a significant role in progress of neural network and deep learning areas. YouTube-8M is such a benchmark dataset for general multi-label video classification. It was created from over 7 million YouTube videos (450,000 hours of video) and includes video labels from a vocabulary of 4716 classes (3.4 labels/video on average). It also comes with pre-extracted audio \& visual features from every second of video (3.2 billion feature vectors in total). 
	
	Google cloud recently released the datasets and organized \textquoteleft Google Cloud \& YouTube-8M Video Understanding Challenge' on Kaggle. Competitors are challenged to develop classification algorithms that assign video-level labels using the new and improved Youtube-8M V2 dataset. 
	
	Inspired by the competition, we started exploration of audio understanding and classification using deep learning algorithms and ensemble methods. We built several baseline predictions according to the benchmark paper \cite{abu2016youtube} and public github tensorflow code. Furthermore, we improved global prediction accuracy (GAP) from base level 77\% to 80.7\% through approaches of ensemble.
	
\end{abstract}

%%%%%%%%% BODY TEXT
\section{Introduction}

There is an English idiom: \textquotedblleft a picture is worth a thousand words". Such theory has been standing in human society for many years, as this is the way our brain functions. With the development of neural network and deep learning, it can be applied to machine as well. In other words, we human beings are able to teach or train computer to recognize objects from pictures and even describe them in our natural language. Thanks Google for organizing this \textquoteleft Google Cloud \& YouTube-8M Video Understanding Challenge', which gives us a wonderful opportunity to test new ideas and implement them with the Google cloud platform. In this paper, we first review the baseline algorithms, and then introduce the innovative ensemble experiments.

%Large-scale datasets have played significant roles in progress of machine learning techniques. YouTube-8M is such a benchmark dataset that contains millions of videos for general multi-label video classification \cite{abu2016youtube}. Motivated by Google Cloud \& YouTube-8M Video Understanding Challenge on Kaggle, we started exploration of audio understanding. In this paper, we first review the baseline algorithms, and then introduce the innovative ensemble experiments.

\section{Data}

There are two types of data, video-level and frame-level features. The data and detailed information are well explained in YouTube-8M dataset webpage (\url{https://research.google.com/youtube8m/download.html}). Both of the video-level and frame-level data are stored as tensorflow.Example protocol buffers, which have been saved as \textquoteleft tfrecord' files.

For each type of the data, it has been split in three sets: train, validate and test. The numbers of observations in each dataset are given in following table.

\begin{tabular}{|c|c|c|c|}
	\hline  &Train  & Validate &Test  \\ 
	\hline Number of Obs & 4,906,660 & 1,401,82 & 700,640  \\ 
	\hline
\end{tabular} 

Video-level data example proto is given in following text format:
\begin{itemize}
	\item 	\textquotedblleft video\_id": an id string;
	\item 	\textquotedblleft labels": a list of integers;
	
	Note: the feature for test set is missing, but given for train and validation datasets.
	\item \textquotedblleft	mean\_rgb": a 1024-dim float list;
	\item \textquotedblleft mean\_audio": a 128-dim float list.
	
\end{itemize}

A few examples of video-level data are given in following table:

\begin{center}
	% \centering
	\resizebox{\linewidth}{!}{
		\begin{tabular}{cccc}
			\hline\hline 
			video\_id (ID) &labels ($y$)&mean\_rbg ($X_1$)&mean\_audio ($X_2$)\\\hline 
			\textquoteleft -09K4OPZSSo'&[66]&
			\begin{tabular}{c}
				[0.11, -0.87, \\
				-0.19, -0.22, \\
				0.23, $\cdots$]
			\end{tabular}
			&
			\begin{tabular}{c}
				[0.89, 1.31, \\
				-0.13, 0.20,\\
				-1.76, $\cdots $]
			\end{tabular}
			\\\hline
			\textquoteleft -0MDly\_IiNM' & [37, 101, 29, 23] &
			\begin{tabular}{c}
				[-0.99, 1.02, \\
				-0.74, 0.09, \\
				0.56, $\cdots $]
			\end{tabular}
			&
			\begin{tabular}{c}
				[-0.66, -1.12, \\
				0.61, -1.37,\\
				-0.01, $\cdots $]
			\end{tabular}\\\hline
			\multicolumn{4}{c}{$\vdots $}
			\\\hline\hline 
	\end{tabular}}
\end{center}

Frame-level data has similar \textquoteleft video\_id' and \textquoteleft labels' features, but \textquoteleft rgb' and \textquoteleft audio' are given in each frame:
\begin{itemize}
	\item \textquotedblleft	video\_id": e.g. \textquoteleft -09K4OPZSSo';
	\item \textquotedblleft labels": e.g. [66];
	\item 	Feature\_list \textquotedblleft rgb": e.g. [[a 1024-dim float list], [a 1024-dim float list],...];
	\item Feature\_list \textquotedblleft audio": e.g. [[a 128-dim float list],[a 12-dim float list],...].
	
	Note: each frame represents one second of the video, which up to 300.
	
\end{itemize}

The data can be represented as $(X_{train},y_{train})$, $(X_{val},y_{val})$ and $(X_{test})$, where $X$ is the information of each video (features), and $y$ is the corresponding labels. In video-level data, $X=(mean\_rgb, mean\_audio)$.

\section{Baseline Approaches}

\cite{abu2016youtube} gave detailed introduction for some baseline approaches, including logistic regression and mixture of experts for video-level data, as well as  frame-level logistic regression, deep bag of frame and long short-term memory models for frame-level data.

Given the video-level representations, we train independent binary classifiers for each label using all the data. Exploiting the structure information between the various labels is left for future work. A key challenge is to train these classifiers at the scale of this dataset. Even with a compact video-level representation for the 6M training videos, it is unfeasible to train batch optimization classifiers, like SVM. Instead, we use online learning algorithms, and use Adagrad to perform model updates on the weight vectors given a small mini-batch of examples (each example is associated with a binary ground-truth value).

\subsection{Models from Video-level Features}

The average values of rgb and audio presentation are extracted from each video for model training. We focus on performance of logistic regression model and mixture of experts(MoE) model, which can be trained within 2 hours on Google cloud platform. 

\subsubsection{Logistic Regression }

Logistic regression computes the weighted entity similarly to linear regression, and obtains the probability by output logistic results \cite{6}. Its cost function is called as log-loss: 
%
%Given $D$ dimensional video-level features, the parameters $\Theta $ of the logistic regression classifier are the entity specific weights $W_e$. During scoring, given $x\in \mathcal{R}^{D+1}$ to be the video-level feature of the test example, the probability of the entity $e$ is given as $p(e|x)=\sigma (W_e^Tx).$ The weights $W_e$ are obtained by minimizing the total log-loss on the training data given as:
\begin{equation}
\lambda||W_e||^2_2+\sum_{i=1}^{N}\mathcal{L}(y_{i,e},\sigma(W_e^Tx_i)),
\end{equation}
where $\sigma(\cdot)$ is the standard logistic, $\sigma(z)=1/(1+exp(-z)).$ The optimal weights can be found with gradient descent algorithm.

\subsubsection{Mixture of Experts (MoE)}
Mixture of experts (MoE) was first proposed by Jacobs and Jordan \cite{jordan1994hierarchical}. %The binary classifier for an entity e is composed of a set of hidden states, or experts, $H_e$. A softmax is typically used to model the probability of choosing each expert. Given an expert, we can use a sigmoid to model the existence of the entity. Thus, the final probability for entity $e$'s existence is $p(e|x) = \sum_{h\in H_e} p(h|x)\sigma(u_h^Tx)$, where $p(h|x)$ is a softmax over $|H_e|+1$ states. In other words, $p(h|x)=\frac{exp(w_h^Tx)}{1+\sum_{h' \in H_e} exp(W_{h'}^Tx)}$. The last, $(|H_e| + 1)^{th}$, state is a dummy state that always results in the non-existence of the entity. Denote $p_{y|x} = p(y = 1|x), p_{h|x} = p(h|x)$ and $p_h = p(y = 1|x, h)$. 
Given a set of training examples $(x_i, g_i), i=1...N$ for a binary classifier, where $x_i$ is the feature vector and $g_i \in [0,1]$ is the ground-truth, let $\mathcal{L}(p_i,g_i)$ be the log-loss between the predicted probability and the ground-truth: 
\begin{equation}
\mathcal{L}(p, g) = - g \log p - (1 - g) \log(1 - p).
\end{equation}
Probability of each entity is calculated from a softmax distribution of a set of hidden states. Cost function over all the datasets is log-loss. 

Video-level models trained with RGB feature achieve 70\% - 74\% precision accuracy. Adding audio feature raises the score to 78\% precision accuracy. We also tested including validation dataset as part of training datasets. With this larger training set, prediction accuracy on tests set is 2.5\% higher than using original training sets only for model fitting. In other words, we used 90\% of YouTube-8M datasets to obtain better model parameters.% without bringing in over-fitting effect.

%We could directly write the derivative of L py|x, g with respect to the softmax weight wh and the logistic weight uh as $\partial L py|x$, g ∂wh = x ph|x py|h,x − py|x  py|x − g  py|x(1 − py|x) , (6) ∂L py|x, g ∂uh = x ph|xpy|h,x(1 − py|h,x) py|x − g  py|x(1 − py|x) . (7) 
%We use Adagrad with a learning rate of 1.0 and batch size of 32 to learn the weights. Since we are training independent classifiers for each label, the work is distributed across multiple machines. 
%
%For MoE models, we experimented with varying number of mixtures (1, 2, 4), and found that performance increases by 0.5\%-1\% on all metrics as we go from 1 to 2, and then to 4 mixtures, but the number of model parameters correspondingly increases by 2 or 4 times. We chose 2 mixtures as a good compromise and report numbers with the 2-mixture MoE model for all datasets.

\subsection{Models from Frame-level Features}
Videos are decoded at one frame-per-second to extract frame-level presentations. In our experiments, most frame-level models achieve similar performance as video-level models.

\subsubsection{Frame-level Logistic Model}
The frame-level features in the training dataset are obtained by randomly sampling 20 frames in each video. Frames from the same video are all assigned the ground-truth of the corresponding video. There are totally about 120 million frames. So the frame-level features are:
$$(x_i,y_i^\epsilon), \epsilon=1,...,4800, i=1,\cdots,120M$$
where $x_i \in \mathcal{R}^{1024}$ and $y_i^\epsilon\in\{0,1\}.$

The logistic models are trained in \textquotedblleft one-vs-all" sense; hence there are totally 4800 models. For inference on test data, we compute the probability of existence of label $e$ in each video $\nu $ as follows
\begin{equation}
p_\nu (e|x_{1:F_\nu }^{\nu })=\frac{1}{F_\nu}\sum_{j=1}^{F_\nu}p_\nu(e|x_j^{\nu}), j=1,\cdots,F_\nu,
\end{equation}
where $F_\nu$ is the number of frames in a video. That is, video-level probabilities are obtained by simply averaging frame-level probabilities.

\subsubsection{Deep Bag of Frame (DBoF) Pooling}
Deep bag of frame model is a convolutional neural network. The main idea is to design two layers in the convolutional part. In the first layer, the up-projection layer, the weights are still applied on frames, although all selected frames share the same parameter. The second layer is pooling the previous layer into video level. The approach enjoys the computational benefits of CNN, while at the same time the weights on the up-projection layer can still provide a strong representation of input features on frame level.

For implementation and test, more features and input data can slightly improve the results. For example, if we combined training and validate data, the score will be improved. When we add both features (RGB + audio), the result is boosted by around 0.4\%. The computing cost is quite low compared to other frame level models. It took 36 hours using one single GPU for training data. The benchmark model is not well implemented for parallel computing in the prediction stage. Using more GPUs doesn't boost training speed. We tested 4 GPU, and the total time is only reduced by 10\%. 

\subsubsection{Long Short-Term Memory (LSTM) Model Trained from Frame-Level Features}

\begin{figure}[!htbp]
	\centering
	\includegraphics[width=0.7\linewidth]{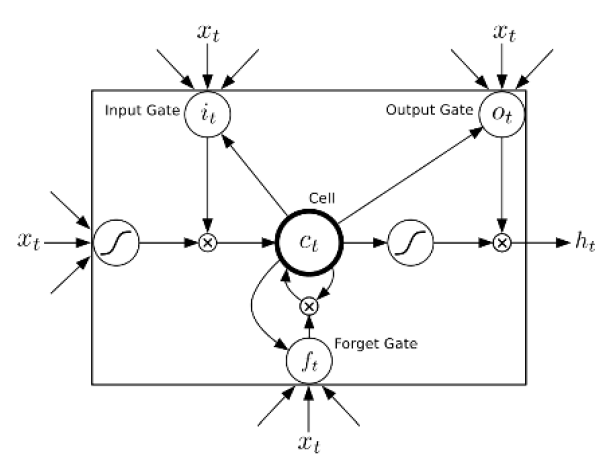}
	\caption{}
	\label{fig:LSTM_plot}
\end{figure}

Long short-term memory (LSTM) \cite{hochreiter1997long} is a recurrent neural network (RNN) architecture. In contrast to conventional RNN, LSTM uses memory cells to store, modify, and access internal state, allowing it to better discover long-range temporal relationships. As shown in Figure \ref{fig:LSTM_plot} (\cite{yue2015beyond}) the LSTM cell stores a single floating point value and it maintained the value unless it is added to by the input gate or diminished by the forget gate. The emission of the memory value from the LSTM cell is controlled by the output gate.

The hidden layer $H$ of the LSTM is computed as follows \cite{yue2015beyond}:

\begin{eqnarray}
i_t &=& \sigma (W_{xi}X_{t} +W_{hi}h_{t-1} +W_{ci}c_{t-1} + b_{i})\\
f_t &=& \sigma (W_{xf}X_{t} +W_{hf}h_{t-1} +W_{cf} c_{t-1} + b_f )\\
c_t &=& f_{t}c_{t-1} + i_t \tanh(W_{xc}X_t +W_{hc}h_{t-1} + b_c)\\
o_t &=& \sigma (W_{xo}x_t +W_{ho}h_{t-1} +W_{co}c_t + b_o)	\\	
h_t &=& o_t \tanh(c_t) 
\end{eqnarray}
where $x$ denote the input, $W$ denote weight matrices (e.g. the subscript hi is the hidden-input weight matrix), $b$ terms denote bias vectors, $\sigma $ is the logistic sigmoid function, and $i, f, o$, and $c$ are respectively the input gate, forget gate, output gate, and cell activation vectors. 

In the current project, the LSTM model was build followed a similar approach to \cite{yue2015beyond}. Provided best performance on the validation set, 2 stacked LSTM layers with 1024 hidden units and 60 unrolling iterations were used \cite{abu2016youtube}.

\section{Ensemble Approaches}
Several predictions on the base level have been generated. Most of the models perform reasonably well, and aggregating the predictions even better results. It is known as ensemble learning to combine the base level models and train a second level to improve the prediction.

There are several approaches of ensemble learning, such as blending, averaging, bagging, voting, etc. Where blending (\cite{bigchaos}) is a powerful method for model ensemble. Averaging, the simple solution, works well, and is also applied in this case. Bagging (short for bootstrap aggregating) is to train the same models on different random subsets of the training sets, and aggregating the results. The improvement from bagging could be limited.  Majority voting is not appropriate in this case, since the outputs here are the confidence level probabilities instead of label ids. 

\subsection{Blending}
A detailed introduction for popular kaggle ensembling methods are given in \url{https://mlwave.com/kaggle-ensembling-guide}. Define $f_1, f_2, \cdots $ be different classification models, like logistic and MoE in this case.

The general idea for blending method is given as follows:  
\begin{itemize}
	\item Create a small holdout set, like 10\% of the train set. 
	\item Build the prediction model with rest of the train set. 
	\item Train the stacker model in this holdout set only.
\end{itemize}

Training the blend model for the YouTube-8M data requires large computational memories.
Google cloud platform provides sufficient memory and computing resources for blending.
We tried this blending method in video-level logistic and MoE base models with validation set as the holdout set.

\begin{itemize}
	\item Build the logistic and MoE model on video-level train data, $y_{train}\sim X_{train}.$
	
	\item Do the inference and print the output prediction on validation set ($\hat{y}_{val}^{logistic}, \hat{y}_{val}^{moe}$) and test set ($\hat{y}_{test}^{logistic}, \hat{y}_{test}^{moe}$).
	
	\item Predictions on dataset $p$ ($p=$ val or test) with model $q$ ($q=$ logistic or MoE), $\hat{y}_p^q$, gives top 20 predictions and their probabilities, e.g.,
	\begin{center}
		\resizebox{\linewidth}{!}{
			\begin{tabular}{ll}
				\hline\hline 
				VideoId & LabelConfidencePairs \\\hline 
				100011194 &
				\begin{tabular}{l}
					1 0.991708 4 0.830637 1833 0.781667 \\
					2292 0.730538 297 0.718730 3547 0.465280 \\
					34 0.396639 1511 0.371649 2 0.351788 \\
					0 0.303522 92 0.169908 933 0.164513 \\
					198 0.145657 202 0.143494 658 0.106776 \\
					74 0.089043 167 0.088266 33 0.052943 \\
					332 0.049101 360 0.045714
				\end{tabular}
				\\\hline 
				100253546 &
				\begin{tabular}{l}
					77 0.996484 21 0.987201 142 0.971881 \\
					59 0.931193 112 0.817585 0 0.445608 \\
					8 0.112624 11 0.100307 17 0.025623 \\
					262 0.021074 1 0.020778 312 0.020060 \\
					75 0.017796 57 0.011925 60 0.005532 \\
					67 0.004512 69 0.004346 575 0.004044 \\
					3960 0.003965 710 0.003961 
				\end{tabular}\\\hline 
				\multicolumn{2}{c}{$\vdots $}
				\\\hline\hline 
		\end{tabular}}
	\end{center}
	
	In order to apply second stage stacker model. We define stacking new feature $X_{p}^{q*}$ to be a 4716-dimension vector. Each dimension represents a label with corresponding probability. The stacking new feature has default value 0, and the 20 estimated probabilities are defined. For simplicity, if there are 3 estimated label with probabilities, the new defined feature is given as following:
	
	$$(\hat{y}_{p}^{q})_i=[1\quad 0.99\quad  4\quad  0.83 \quad 5\quad  0.78]$$
	$\Rightarrow$
	$$(X_{p}^{q*})_{i}=(0.99, 0,0,0.83,0.78,0,\cdots,0)$$
	%
	%$$(X_{p}^{q*})_{ij}=\left[
	%\begin{array}{ll}
	%\hat{p},& (\hat{y}_{p}^{q})_i \quad has \quad prediction \quad on \quad label \quad j\\
	%0, & o.w.
	%\end{array}
	%\right.$$  
	where $i=1, \cdots, 1401828$ if $p=$val, $i=1, \cdots, 700640$ if $p=$test. Write the new defined stacking feature to be $X_{val}^{*}=(X_{val}^{logistic *}, X_{val}^{moe *})$ and $X_{test}^{*}=(X_{test}^{logistic *}, X_{test}^{moe *})$.
	
	\item Stacker model is hard to run in local machine with new defined feature $X_{val}^{*}$,  since it costs too much memories. Hence, we save the new defined data as tfrecord file, so that the model can be run in Google cloud. The new defined data example proto is given as following: 
	
	\begin{itemize}
		\item[a.] \textquotedblleft video\_id"
		\item[b.] \textquotedblleft labels"
		\item[c.] \textquotedblleft logistic\_newfeature": float array of length 4716
		\item[d.] \textquotedblleft moe\_newfeature": float array of length 4716
	\end{itemize}

	\item Train the stacker model in the validation new feature $X_{val}^*$,  $y_{val}\sim X_{val}^*.$  In this case, we use only logistic and MoE as stacker model, since these two are well-defined in th package. Other models need to be defined if desire to use.
	
	\item Do the inference on the test set $X_{test}^*$, hence final prediction $\hat{y}_{test}^*.$
\end{itemize}

Note since the new defined feature has length 4716 instead of 1024, train.py and test.py in Google cloud script need some small correction to apply this method. Defined feature\_names and feature\_sizes need to match with new defined feature name and corresponding length. By adding the new features, the score has been improved from 0.760 to 0.775, see Table \ref{tab:predictions}.

\subsection{Averaging}
The idea of averaging is to generate a smooth separation between different predictors and reduce over-fitting. For each video, if one label is predicted several times among these models, the mean value of these confidence level values is used as the final prediction. On the other hand, if a label id is rarely predicted among the base models, the final prediction is calculated using the sum of these confidence level values divided by the number of base models, eventually lower the confidence level. On a computer with 8G physical memories, 30G virtual memory, the averaging process for the whole test dataset, which includes 700640 YouTube video records, can be finished within 15 minutes, which is quite efficient considering the size of files.

Several strategies are tested for the averaging: 

Strategy A: 3 frame level models, 2 video level models, 2 blending models (video level logistic model blending with MoE prediction, video level MoE model blending with MoE prediction), the GAP score is 0.80396 after averaging, higher than the score of any individual base predictions.

Strategy B:  3 frame level models, 2 video level models, 2 blending models (video level logistic model blending with both logistic and MoE prediction, video level MoE model blending with both logistic and MoE prediction). The basement of the 2 blending predictions are the 2 video level models, thus the 2 blending models are highly correlated with the 2 video level models. The video level logistic model is a weaker predictor than video level MoE predictor. The usage of logistic prediction blended models increases the weight of this weaker predictor. Therefore, the GAP score is 0.80380 after averaging, slightly lower than Strategy A.

Strategy C:  the base models are the same as in strategy A, except that the deep bag of frame pooling frame model prediction is replaced by the one generated after fine-tuning the hyper-parameters. Including the individual model with better performance, the score after averaging is improved to 0.80424.

Strategy D: the base models in strategy D is the same as strategy C, but the weight of the blending model (video level MoE model blending with both logistic and MoE prediction) is increased to 2. Among all the base models, the GAP score of the MoE blending model is the highest. The score rises to 0.80692 with the weighted average strategy.

Strategy E: similar to strategy D, the weight of the MoE blending model is 2. The logistic model blending with logistic plus MoE prediction is replaced by the logistic model blending only with MoE prediction, which reduces the logistic components compared to strategy D. This approach yields the best GAP score, 0.80695. 

Such method boosts the final result by 7-8\%, quite remarkable considering the simplicity nature. Table \ref{tab:predictions} shows results of all base models, where the scores are calculate using GAP metrics on Kaggle platform.

\begin{figure}
	\centering
	\includegraphics[width=\linewidth]{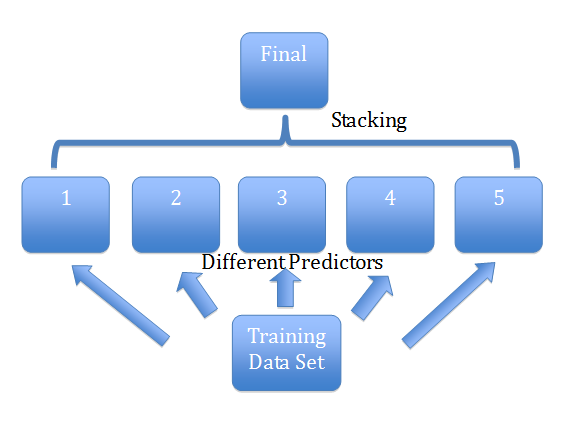}
	\caption{Stacking Diverse Predictors Flow Chart}
	\label{fig:stacking_plot}
\end{figure}

\begin{table}
	\caption{Predictions from Individual Base Models}
	\label{tab:predictions}
	\centering
	\resizebox{\linewidth}{!}{
		\begin{tabular}{|c|c|c|c|}
			\hline Models & Features &Datasets for model fitting  &Score  \\ \hline 
			\begin{tabular}{c}
				Frame Level\\
				LSTM Model
			\end{tabular}
			&rgb&validate&0.7457\\\hline 
			\begin{tabular}{c}
				Frame Level\\
				Deep Bag of \\
				Pooling Model I
			\end{tabular}
			&audio&train+validate&0.77 \\\hline 
			\begin{tabular}{c}
				Frame Level \\
				Deep Bag of \\
				Pooling Model II
			\end{tabular}
			&audio&train&0.767\\\hline 
			\begin{tabular}{c}
				Video Level \\
				Logistic Model
			\end{tabular}
			&audio/rgb&train+validate&0.76036\\\hline
			\begin{tabular}{c}
				Video Level \\
				Mixture of \\
				Experts Model
			\end{tabular}
			&audio/rgb&train+validate&0.78453\\\hline
			\begin{tabular}{c}
				Video Level \\
				Logistic Model \\
				Blending I
			\end{tabular}
			&audio/rgb/logistic/moe&validate&0.77518\\\hline
			\begin{tabular}{c}
				Video Level \\
				Logistic Model \\
				Blending II
			\end{tabular}
			&audio/rgb/moe&validate&0.76873\\
			\hline
			\begin{tabular}{c}
				Video Level \\
				Mixture of \\
				Experts Model \\
				Blending
			\end{tabular}
			&audio/rgb/logistic/moe&validate&0.78617\\
			\hline 
		\end{tabular} 
	}
\end{table}

\section{Conclusion}

Utilizing the open resource of Youtube-8M train and evaluation datasets, we trained baseline models in a fast-track fashion. Video-level representations are trained with a) logistic models; b) MoE model, while frame-level features are trained with a) LSTM model; b) Dbof model; c) frame-level logistic model. We also demonstrated the efficiency of blending and averaging to improve the accuracy of prediction. Blending plays a key role to raise the performance of baseline. We are able to train the blender model on Google cloud, where the computer resource can digest thousands of features. The averaging solution, in addition, aggregates the wisdom from all predictions with low computer cost.

{\small
	\bibliographystyle{ieee}
	
	%\bibliography{youtube8m_bib}
}

\end{document}